\documentclass[12pt]{article}
\usepackage{comment}
\usepackage{amsmath}
\usepackage{amssymb}
\usepackage{graphicx}
\usepackage{here}
\usepackage{mathtools}
\usepackage{ascmac}
\usepackage{subcaption}
\usepackage{url}
\usepackage{mathtools}
\usepackage[sort&compress, numbers, merge]{natbib}
\usepackage{physics}
\usepackage{todonotes}

\usepackage[colorlinks=true, linkcolor=blue, citecolor=blue, urlcolor=black]{hyperref}

\begin{document}

\def\ps{\mathbf{p}}
\def\PS{\mathbf{P}}

\baselineskip 0.6cm

\def\simgt{\mathrel{\lower2.5pt\vbox{\lineskip=0pt\baselineskip=0pt
           \hbox{$>$}\hbox{$\sim$}}}}
\def\simlt{\mathrel{\lower2.5pt\vbox{\lineskip=0pt\baselineskip=0pt
           \hbox{$<$}\hbox{$\sim$}}}}
\def\simprop{\mathrel{\lower3.0pt\vbox{\lineskip=1.0pt\baselineskip=0pt
             \hbox{$\propto$}\hbox{$\sim$}}}}
\def\tr{\mathop{\rm tr}}
\def\SU{\mathop{\rm SU}}

\begin{titlepage}

\begin{flushright}
\end{flushright}
~

\vskip 1.5cm

\begin{center}

{\Large \bf
Learning principle and mathematical realization of the learning mechanism in the brain}

\vskip 2cm
{\large
Taisuke Katayose\footnote{\texttt{tai.katayose@gmail.com}}
}

\vskip 1.0cm
{\it

}

\vskip 3.5cm
\abstract{While deep learning has achieved remarkable success, there is no clear explanation about why it works so well. In order to discuss this question quantitatively, we need a mathematical framework that explains what learning is in the first place. After several considerations, we succeeded in constructing a mathematical framework that can provide a unified understanding of all types of learning, including deep learning and learning in the brain. We call it learning principle, and it follows that all learning is equivalent to estimating the probability of input data. 
We not only derived this principle, but also mentioned its application to actual machine learning models. For example, we found that conventional supervised learning is equivalent to estimating conditional probabilities, and succeeded in making supervised learning more effective and generalized. We also proposed a new method of defining the values of estimated probability using differentiation, and showed that unsupervised learning can be performed on arbitrary dataset without any prior knowledge. Namely, this method is a general-purpose machine learning in the true sense. Moreover, we succeeded in describing the learning mechanism in the brain by considering the time evolution of a fully or partially connected model and applying this new method. The learning principle provides solutions to many unsolved problems in deep learning and cognitive neuroscience.
}

\end{center}

\end{titlepage}

\tableofcontents
\newpage

\section{Introduction}
\label{sec: Introduction}
Since the usefulness of deep neural networks was demonstrated\,\cite{hinton06}, deep learning has made great progress and is being applied to a variety of fields. In particular, the performance of models that specialize in specific data has been astounding. For example, models that incorporate CNNs\,\cite{726791} for image recognition and Attention mechanisms\,\cite{vaswani2023attention} for language processing have shocked the world with their high performance. Then, why has deep learning been so successful? Currently, we are only using what has been successful as an ad-hoc measure, and there is no theoretical guarantee that deep learning will always work. For the further development of this field, it is necessary to have an intrinsic understanding about learning itself, which can give clear answers about why deep learning is successful.

There is a rough explanation that deeper layers make it possible to extract complex features, which is the reason for the success of deep learning. However, there is no mathematically rigorous definition of features, and the question of ``why deep learning works" is just replaced by the question of ``why features can be extracted". Rather than being satisfied with such a qualitative explanation, it must be evaluated using quantitative methods to gain an essential understanding. So let us reconsider in a quantitative way the most basic concept of what is the success in learning. If we evaluate the supervised learning, learning is considered to be successful when it can make predictions close to the teacher labels. Then, where is the guarantee that these teacher labels are actually correct? For example, teacher labels for tasks such as image classification or language translation are manually assigned by humans, but there is no mathematical necessity for this. Teacher labels are created as a result of some kind of human learning process, and the discussion must start from the question of why it is correct. Therefore, in order to mathematically and quantitatively evaluate machine learning, it is necessary to have a framework in which even human learning, i.e., learning by the human brain, can be discussed in a unified manner. Such a framework is called the learning principle, and the purpose of this paper is to clarify it. The first half of this paper up to Chapter\,\ref{sec: Definition of learning principle} describes the derivation of the learning principle, and the second half starting from Chapter\,\ref{sec: Relation to other loss functions} describes the application of the learning principle.

\section{Philosophy of learning principle}
\label{sec: Principle of learning}
In this chapter, we describe the underlying idea behind the derivation of the learning principle. The details of the mathematical calculations along the way are given in Chapter\,\ref{sec: Derivation of the loss function}, and the full definition of the learning principle is given in Chapter\,\ref{sec: Definition of learning principle}.

\subsection{Three essential elements of learning}
Our goal is to derive a learning principle that can uniformly describe all learning, including machine learning and learning in the human brain. To this end, we will consider the elements that are common to all learning.

First, in any type of learning, there are targets to be optimized. For example, in the case of deep learning, parameters such as weights and biases are the targets of optimization, and in the case of the brain, this corresponds to the way neurons are connected to each other. Neural networks and the brain itself can be regarded as structures that include the targets of these optimizations, and in the following, such structures will be referred to as models, borrowing the terminology from deep learning.

Second, any type of learning requires input data. In the case of machine learning, this is the input itself to the model, and in the case of the brain, it is the information from the five senses.

Third, in any type of learning, an optimization strategy must be defined. In the case of machine learning, basically some objective function is defined in advance, and the model is optimized by minimizing or maximizing its value. In the case of the brain, such an objective function is not clear, but we will proceed on the assumption that something equivalent exists. In the following, borrowing the terminology of deep learning, the objective function will be referred to as a loss function, and optimization will be performed by minimizing the loss function.

Based on the above considerations, three elements are essential for learning: a model, input data, and a loss function. Conversely, learning can be performed if at least these elements are defined. In what follows, we will proceed with an abstract discussion about the three essential elements, without considering concrete aspects such as the internal structure of the model, the type of input data, and the computational method used to minimize the loss function.

\subsection{Thought experiment for ideal case}
In order to understand the essence of learning, we will conduct a thought experiment about learning in an ideal situation. The ideal situation here is one in which the model has a universal approximation property\,\cite{HORNIK1989359} and an infinite amount of input data and computational resources are available. Regarding to the loss function, any function can be used. Let us briefly explain universal approximation property here. Any model can be regarded as a function that receives input and returns some kind of output, and in the following, such a function will be referred to as a model function. If a model function can approximate any function by changing its internal parameters, the model is said to have a universal approximation property.

Let us go back to the topic and summarize what kind of results can be obtained if learning is actually carried out under ideal conditions. Since we have unlimited input data and computational resources, we can optimize the model as many times as we want, eventually arriving at a solution that minimizes the loss function. Also, since it is assumed that the model has a universal approximation property, the solution here is a model function that truly minimizes the loss function. Below, when we use the word solution, we will refer to such a model function.

The important thing is that there is only one solution that truly minimizes the loss function. If a model has a universal approximation property, the same model function will ultimately be obtained regardless of its internal structure. That is, the solution depends only on the input dataset and loss function, not on the internal structure of the model. This is a very important consequence, showing that the details of the model are irrelevant when considering the learning principle.

\subsection{The solution which minimizes the loss function}
Now that we know that the details of the model are unrelated to the learning principle, the next thing to consider is the relationship among the input data, the loss function, and the solution that minimizes it. First, we consider supervised learning in deep learning as an example. In supervised learning, an input dataset and corresponding teacher labels are given, and the goal is to construct a model that predicts the teacher label from the input data as accurately as possible. Namely, what is expected as a solution is a model function that predicts correct teacher labels for any input data. The loss function used for learning must be the function that is minimized by such a model function. An example of a loss function that satisfies this condition is the mean square error. As a matter of fact, the value of the mean square error is minimized, when the model function always predicts the correct teacher label.

Next, a more general case will be explained. Since we want to find a framework that can understand all learning in a unified manner, we will consider the most general case, that is, unsupervised learning without any prior knowledge of the input data. What we wanted to emphasize in the previous example of supervised learning is that we first assumed the desired solution and then considered the loss function to obtain it. So, in the case of unsupervised learning, what kind of model function should be assumed as a solution? Also, assuming a solution, how should we define the loss function to obtain it? These questions will be considered in the next section.

\subsection{Probability of the input}
It is necessary to assume something as a solution, but what kind of information can be extracted with unsupervised learning without prior knowledge? Since we are discussing the most general case here, such information must be definable for any type of input data. The answer to this question is the probability itself that the input data has. We will give an example to explain for ease of understanding. Let us consider the case of learning a large amount of image data represented by black and white dots, and assume that the input is given as binary data of 0 or 1. Of course, we do not use the information that this data represents image data. If the image data is 100 pixels, there are a total of $2^{100}$ possible patterns. If infinite input data is given, even though there are such a large number of patterns, the exact same data will appear repeatedly. At this time, there should be patterns with high and low frequencies of appearance, and this is exactly what we described earlier as the probability that the input data has. There is a probability distribution for any input dataset, and each input has a unique probability. This probability is the only information that is always associated with any type of input data, regardless of its format.

\subsection{Brief summary of learning principle}

From the above discussion, learning is estimating the probability of input data, and the solution must be the model function that returns the true probability of input data. The loss function must be a function which is minimized by such a model function. In Chapter\,\ref{sec: Derivation of the loss function}, we mathematically derive a loss function which satisfies this condition and show that normalization of the estimated probability plays an important role. The complete form of the learning principle is given in Chapter\,\ref{sec: Definition of learning principle}.

\section{Derivation of the loss function}
\label{sec: Derivation of the loss function}
In this chapter, we derive the loss function for learning principle. We consider two patterns where we want to estimate the probability of input itself or conditional probability defined through input. 

\subsection{General case}
\label{sec: General case}
In this section, we focus only on the loss function and the model function without considering the internal structure of the model. Let us consider a model that has a universal approximation property and returns a value $\Phi(\vb{x})$ for input $\vb{x}$\footnote{In this paper, italic characters such as $a$, $b$, $x$ and $y$ are defined as scalar values, bold characters such as $\vb{a}$, $\vb{b}$, $\vb{x}$ and $\vb{y}$ as vectors, and bold characters with a tilde such as $\tilde{\vb{\Sigma}}$ and $\tilde{\vb{W}}$ as matrices.}. Here, $\Phi(\vb{x})$ is the model function to be optimized as learning proceeds, and we want $\Phi(\vb{x})$ to represent the estimation of the probability of input. The loss function must be minimized when $\Phi(\vb{x})=P(\vb{x})$ is satisfied where $P(\vb{x})$ is the probability the input $\vb{x}$ has. Let us assume $L(\Phi(\vb{x}))$ is such a loss function and consider the condition under which it is minimized. Minimizing the loss function means minimizing its expected value defined as
\begin{equation}
\label{eq: expected value of loss (General case)}
    E(L(\Phi)) = \sum_{\vb{x}\in \mathrm{All}}P(\vb{x}) L(\Phi(\vb{x})) \, ,
\end{equation}
if $\vb{x}$ takes continuous values, then 
\begin{equation}
\label{eq: expected value of loss (General case))}
    E(L(\Phi)) = \int_{\vb{x} \in \mathrm{All}}d\vb{x}P(\vb{x}) L(\Phi(\vb{x})) \, .
\end{equation}
Next, let us consider the solution for $\Phi(\vb{x})$ that minimizes this expectation value. However, if we try to just minimize it, we will find that every $\Phi(\vb{x})$ approaches to the same value which is the minimum point of $L(\Phi)$. To avoid this situation, the following normalization condition is imposed.
If $\vb{x}$ takes discrete values, then
\begin{equation}
\label{eq: norm discrete (General case)}
    \sum_{\vb{x}\in \mathrm{All}}\Phi(\vb{x})  = 1 \, ,
\end{equation}
and if $\vb{x}$ takes continuous values, then
\begin{equation}
\label{eq: norm continuous (General case)}
    \int_{\vb{x}\in \mathrm{All}}d\vb{x}\,\Phi(\vb{x})  = 1 \, .
\end{equation}
This normalization condition is reasonable thinking that we want $\Phi(\vb{x})$ to approach $P(\vb{x})$. We then use variational method to find the condition under which the expectation value of the loss function takes minimum. First, we choose any two points $\vb{x}$ and $\vb{x}^\prime$, and take the variation there as 
\begin{equation}
\label{eq: variation x xp (General case)}
\begin{split}
    \Phi(\vb{x}) &\to \Phi(\vb{x}) + \epsilon  \, ,\\
    \Phi(\vb{x}^{\,\prime}) &\to \Phi(\vb{x}^{\,\prime}) - \epsilon  \, ,
\end{split}
\end{equation}
where $\epsilon$ is infinitesimally small value. These variations do not disturb the condition in Eq.\,(\ref{eq: norm discrete (General case)}) or (\ref{eq: norm continuous (General case)}). The variation of $E(L(\Phi))$ is calculated as 
\begin{equation}
\label{eq: variation e (General case)}
\begin{split}
    \delta E(L(\Phi)) &= \left[P(\vb{x}) \frac{\partial L(\Phi(\vb{x}))}{\partial \Phi(\vb{x})} - P(\vb{x}^{\,\prime}) \frac{\partial L(\Phi(\vb{x}^{\,\prime}))}{\partial \Phi(\vb{x}^{\,\prime})}\right]\epsilon \, .
\end{split}
\end{equation}
At the minimum of $E(L(\Phi))$, this equation becomes 0 and $\Phi(\vb{x}) = P(\vb{x})$ also should be satisfied. In order to achieve this, the first term and the second term in the right hand side must be the same constant independent of $\vb{x}$, when $\Phi(\vb{x})=P(\vb{x})$ is satisfied. Then, we conclude
\begin{equation}
\label{eq: const condition (General case)}
\left.P(\vb{x}) \frac{\partial L(\Phi(\vb{x}))}{\partial \Phi(\vb{x})}\right|_{\Phi(\vb{x}) = P(\vb{x})} = const. \, ,
\end{equation}
which leads
\begin{equation}
\label{eq: loss c1 c2 (General case)}
L(\Phi(\vb{x})) = c_1 \log{(\Phi(\vb{x}))} + c_2\, ,
\end{equation}
where $c_1$ and $c_2$ are arbitrary constants. Considering $E(L(\Phi))$ takes minimum not maximum, $c_1$ must be negative. We can set $c_1 = -1$ and $c_2 = 0$ for Eq.\,(\ref{eq: loss c1 c2 (General case)}) without the loss of generality, and the loss function is written as
\begin{equation}
\label{eq: log loss (General case)}
L(\Phi(\vb{x})) = - \log{(\Phi(\vb{x}))}\, .
\end{equation}
This is the loss function we were looking for. Summarizing the discussion above, we obtain the relation as 
\begin{equation}
\label{eq: relation (General case)}
    \big[E(-\log(\Phi(\vb{x}))) \to \min \big] \Longleftrightarrow   \big[ \Phi(\vb{x}) \to P(\vb{x}) \big] \, .
\end{equation}
From this relation, $\Phi(\vb{x})$ is considered as the estimation of $P(\vb{x})$ by the machine learning model, and it approaches the true value of $P(\vb{x})$ after sufficient learning. At this time, the loss function defined in Eq.\,(\ref{eq: log loss (General case)}) is nothing more than the estimation of the self-information that the input $\vb{x}$ has. We note that this loss function works only under the normalization condition of Eq.\,(\ref{eq: norm discrete (General case)}) or Eq.\,(\ref{eq: norm continuous (General case)}) and when $\Phi(\vb{x})$ is positive. Conversely, as long as $\Phi(\vb{x})$ satisfies these conditions, the $\Phi(\vb{x})$ always works as the estimation of the probability $P(\vb{x})$.

\subsection{Conditional probability}
\label{sec: Conditional probability}
In this section, we derive the loss function to estimate conditional probabilities. Here, we will assume that the input takes discrete values, but the case for continuous values is exactly the same except for changing the summations to integrals in the following. Let us consider that input $\vb{x}$ is composed of two vectors as $\vb{x} = (\vb{a},\vb{b})$, then we can decompose $P(\vb{x})$ as
\begin{equation}
\label{eq: probability decomposition}
   P(\vb{x}) = P(\vb{a},\vb{b}) = P(\vb{a})P(\vb{b}|\vb{a}) \, ,
\end{equation} 
where $P(\vb{b}|\vb{a})$ is the conditional probability of $\vb{b}$ under the condition of $\vb{a}$. Now, our goal is to estimate $P(\vb{b}|\vb{a})$, and we consider a machine learning model which return $\Phi(\vb{x}) = \Phi(\vb{a},\vb{b})$ for the input $\vb{x}$. The expectation value of the loss function is calculated as   
\begin{equation}
\label{eq: expected value of loss (Conditional probability)}
\begin{split}    
    E(L(\Phi)) &= \sum_{\vb{x}\in \mathrm{All}}P(\vb{x})L(\Phi(\vb{x})) \\
    &=  \sum_{\vb{a}\in \mathrm{All}} \sum_{\vb{b}\in \mathrm{All}}P(\vb{a})P(\vb{b}|\vb{a})L(\Phi(\vb{a},\vb{b})) \, .
\end{split}
\end{equation}
We want $\Phi(\vb{a},\vb{b})$ to estimate $P(\vb{b}|\vb{a})$, so we impose the following condition as
\begin{equation}
\label{eq: norm condition (Conditional probability)}
\begin{split}
    \forall \vb{a}:\sum_{\vb{b}\in \mathrm{All}}\Phi(\vb{a},\vb{b}) & = 1 \, .
\end{split}
\end{equation}
To use variational method, we choose any two points $\vb{x} = (\vb{a},\vb{b})$ and $\vb{x}^\prime = (\vb{a},\vb{b}^\prime)$ then take the variation as 
\begin{equation}
\label{eq: variation x xp (Conditional probability)}
\begin{split}
    \Phi(\vb{a},\vb{b}) &\to \Phi(\vb{a},\vb{b}) + \epsilon  \\
    \Phi(\vb{a},\vb{b}^\prime) &\to \Phi(\vb{a},\vb{b}^\prime) - \epsilon  \, ,
\end{split}
\end{equation}
where this variation does not disturb the condition in Eq.\,(\ref{eq: norm condition (Conditional probability)}). Under these variations, the variation of $E(L(\Phi))$ is calculated as
\begin{equation}
\label{eq: variation e (Conditional probability)}
\begin{split}
    \delta E(L(\Phi)) &= P(\vb{a})\left[P(\vb{b}|\vb{a}) \frac{\partial L(\Phi(\vb{a},\vb{b}))}{\partial \Phi(\vb{a},\vb{b})} - P(\vb{b}^\prime|\vb{a}) \frac{\partial L(\Phi(\vb{a},\vb{b}^\prime))}{\partial \Phi(\vb{a},\vb{b}^\prime)}\right]\epsilon \, .
\end{split}
\end{equation}
We want this equation to vanish when $\Phi(\vb{a},\vb{b}) = P(\vb{b}|\vb{a})$ is satisfied for any $\vb{a}$ and $\vb{b}$, so we conclude
\begin{equation}
\label{eq: const condition (Conditional probability)}
\left.P(\vb{b}|\vb{a}) \frac{\partial L(\Phi(\vb{a},\vb{b}))}{\partial \Phi(\vb{a},\vb{b})}\right|_{\Phi(\vb{a},\vb{b}) = P(\vb{b}|\vb{a})} = const. \, ,
\end{equation}
which leads
\begin{equation}
\label{eq: loss c1 c2 (Conditional probability)}
L(\Phi(\vb{a},\vb{b})) = c_1 \log{(\Phi(\vb{a},\vb{b}))} + c_2\, .
\end{equation}
We set $c_1= -1$ and $c_2 = 0$, then the loss function is
\begin{equation}
\label{eq: log loss (Conditional probability)}
    L(\Phi(\vb{a},\vb{b})) = - \log{(\Phi(\vb{a},\vb{b}))} \, .
\end{equation}
Summarizing this discussion, we have the following relation as
\begin{equation}
\label{eq: relation (Conditional probability)}
    \big[E(- \log{(\Phi(\vb{a},\vb{b}))}) \to \min \big] \Longleftrightarrow   \big[ \Phi(\vb{a},\vb{b}) \to P(\vb{b}|\vb{a}) \big] \, .
\end{equation}
The loss function is same as that in previous section and the only difference is the normalization condition as in Eq.\,(\ref{eq: norm discrete (General case)}) and Eq.\,(\ref{eq: norm condition (Conditional probability)}). 

\section{Definition of learning principle}
\label{sec: Definition of learning principle}
\begin{itembox}{Learning principle}
Learning requires three elements: targets of optimization, input data, and an objective function. We call the structure which include targets of optimization as a model, an objective function as a loss function, borrowing the terminology of deep learning. \\

A model can be thought of as a function that receives input and returns some output, and we call it a model function. The model function estimates the probability of the input, and must always take a positive value and satisfy the probability normalization condition that the total sum is 1.\\

The loss function is defined by taking the logarithm of the model function and adding a negative sign, and it has the same form as the self-information formula. Optimization is performed by minimizing this loss function, which ensures that the model function automatically approaches the true probability.
\end{itembox}

This is the learning principle derived in Chapters\,\ref{sec: Principle of learning} and \ref{sec: Derivation of the loss function}. All learning can be understood based on this principle, and as long as this condition is met, it can be assumed as learning. Since the form of the loss function is fixed, what we can modify is how to satisfy the normalization conditions for a model function. Furthermore, as long as the model function is always positive and satisfies the normalization condition, it can be interpreted as an estimated probability, no matter how it is defined. In other words, the model function does not need to be defined in the same way as the output in normal deep learning. The details of this point will become clear in later chapters.

In the subsequent chapters, we will mainly discuss the normalization conditions for the model function. In Chapter\,\ref{sec: Relation to other loss functions}, we discuss how conventional machine learning can be understood based on the learning principle, and show in particular that supervised learning is equivalent to the estimation of conditional probabilities. In Chapter\,\ref{sec: Normalization by differential}, we propose a method to satisfy the normalization condition by defining a model function using differentiation. This method makes it possible to perform unsupervised learning on arbitrary data sets without prior knowledge, making machine learning universal in the true sense. In Chapter\,\ref{sec: Normalization by time evolution}, we propose a method that satisfies the normalization condition by defining a model function based on the time evolution of a fully or partially connected model. This method leads to a completely new concept of a loss function localized in time and space, which allows us to identify our model as a mathematical description of the learning mechanism in the brain. In Chapter\,\ref{sec: Discussion}, we summarize the results and findings brought about by the learning principle and reaffirm its importance and validity.

\section{Learning principle for some problems}
\label{sec: Relation to other loss functions}
In this chapter, we discuss how to apply learning principle for some problems. We will also see how conventional machine learning can be understood in the perspective of learning principle. Before looking at individual problems, let us see how commonly used loss functions can be understood in the context of learning principle. The argument in Chapter\,\ref{sec: Derivation of the loss function} is very general and the relation in Eq.\,(\ref{eq: relation (General case)}) or (\ref{eq: relation (Conditional probability)}) always holds. Rewriting these relationships in terms of the loss function, we get
\begin{equation}
\label{eq: relation 1 (Relation to other loss functions)}
    \big[E(L) \to \min \big] \Longleftrightarrow   \big[ \exp(-L(\vb{x})) \to P(\vb{x}) \big] \, ,
\end{equation}
or
\begin{equation}
\label{eq: relation 2 (Relation to other loss functions)}
    \big[E(L) \to \min \big] \Longleftrightarrow   \big[ \exp(-L(\vb{x})) \to P(\vb{b|a}) \big] \, ,
\end{equation}
where $L$ can be any loss function, for instance mean squared error. Here, $\exp(-L(\vb{x}))$ can be interpreted as the model function and it must satisfy certain normalization condition. We can conclude that using certain loss function $L(\vb{x})$ and minimizing it is equivalent to approximating the probability of the input by $\exp(-L(\vb{x}))$. 

\subsection{Classification problem}
\label{sec: Classification problem}
\begin{figure}[t]
    \centering
    \includegraphics[scale=0.8]{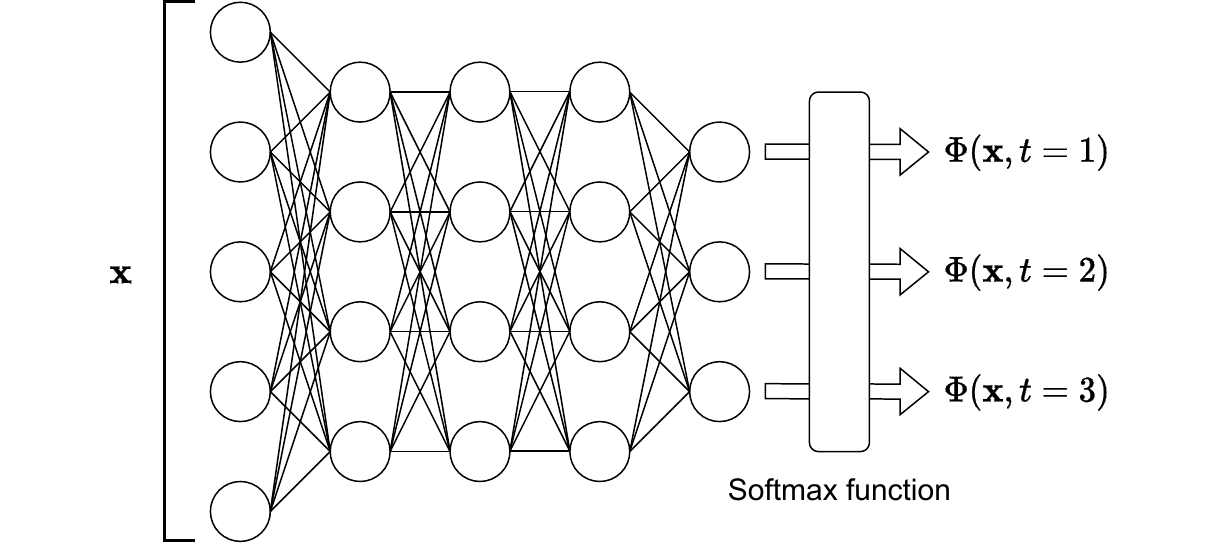}
    \caption{\small \sl The example of the model for classification problem. In this figure, we assume the classification for 3 labels.}
    \label{fig: classification}
\end{figure}

In a classification problem, we are usually given a set of input data $\{\vb{x}_1,\vb{x}_2,\cdots\}$ and a set of teacher labels $\{\vb{t}_1,\vb{t}_2,\cdots\}$. If we regard the set of $(\vb{x},\vb{t})$ as one input data, the discussion in Sec.\,\ref{sec: Conditional probability} can be applied. Namely, it is considered as an unsupervised learning for the dataset  $\{(\vb{x}_1,\vb{t}_1),(\vb{x}_2,\vb{t}_2),\cdots\}$.
Thinking in this way, it becomes clear that the teacher label is just a numerical value included in the input data, and is not an absolute correct answer. In other words, different teacher labels may be attached to the exact same input. Each combination of input data and teacher labels is considered to have a unique probability, and the goal is to clarify that probability. In conventional machine learning, it is assumed that there is a correct answer called the teacher label, and the goal is to create a model that makes predictions close to that answer, but this idea was wrong in the first place.
In a classification problem, the probability distribution we want to get is the conditional probability of the teacher label $\vb{t}$ when the input $\vb{x}$ is given, which is written as $P(\vb{t}|\vb{x})$. In this case, we consider the model function $\Phi(\vb{x},\vb{t})$ and we define the loss function as 
\begin{equation}
\label{eq: log loss (Classification ploblem)}
L(\Phi(\vb{x},\vb{t})) = -\log{(\Phi(\vb{x},\vb{t}))} \, ,
\end{equation}
under the normalization condition written as
\begin{equation}
\label{eq: norm condition (Classification problem)}
\begin{split}
    \forall \vb{x}: \sum_{\vb{t} \in \mathrm{All}}\Phi(\vb{x},\vb{t}) & = 1 \, .
\end{split}
\end{equation}
This normalization condition is accomplished by using the following model. The model is a usual neural network as shown in Fig.\ref{fig: classification} whose input layer takes $\vb{x}$ as an input and output layer has as many nodes as the kinds of the label. The sum of the values of the nodes in output layer is normalized by softmax function, and $\Phi(\vb{x},\vb{t})$ is defined as the value of the corresponding node in output layer. In this case, the loss function becomes exactly the same as the cross-entropy with one-hot vector. The same loss function has been used in conventional deep learning, but its true meaning is not to reduce the error with the teacher label, but to estimate the conditional probability $P(\vb{t}|\vb{x})$. 

\subsection{Regression problem}
\label{sec: Regression problem}
The regression problem differs from the classification problem in that the teacher labels are continuous. We consider the model function $\Phi(\vb{x},\vb{t})$, and define the loss function as
\begin{equation}
\label{eq: log loss (Regression problem)}
L(\Phi(\vb{x},\vb{t})) = -\log{(\Phi(\vb{x},\vb{t}))} \,  ,
\end{equation}
under the normalization condition written as
\begin{equation}
\label{eq: norm condition (Regression problem)}
\begin{split}
    \forall \vb{x}: \int d\vb{t}\,\Phi(\vb{x},\vb{t}) & = 1 \, .
\end{split}
\end{equation}
Eq.\,(\ref{eq: norm condition (Regression problem)}) contains integration, and the model discussed in Sec.\,\ref{sec: Classification problem} cannot be used. In this section, we discuss the prescription to satisfy this normalization condition by considering some approximations and assumptions. The general prescription without any approximations or assumptions will be discussed in Sec.\,\ref{sec: Regression problem for general case}.

\begin{figure}[t]
    \centering
    \includegraphics[scale=0.8]{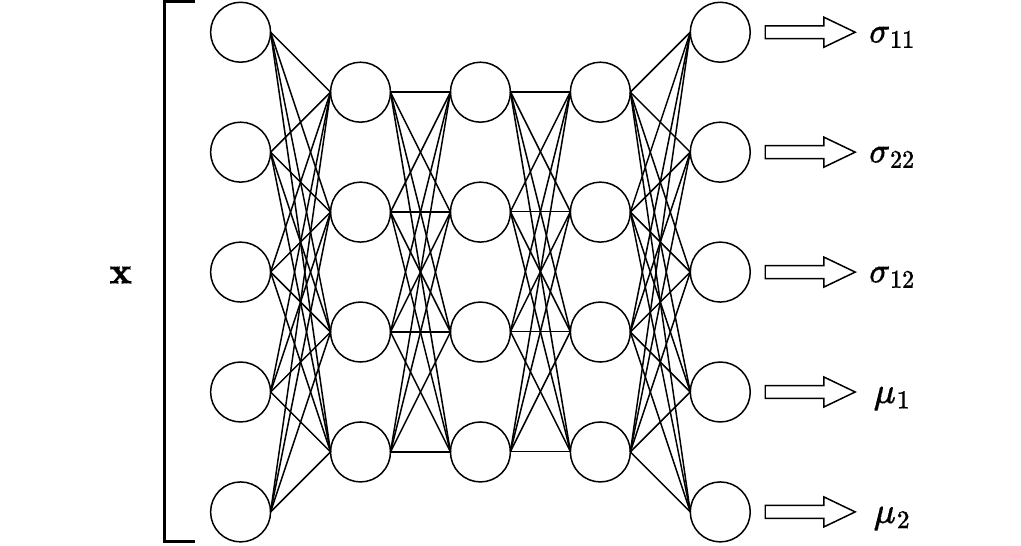}
    \caption{\small \sl The example of the model for regression problem. In this figure, we assume the regression into 2 parameters.}
    \label{fig: regression}
\end{figure} 

Let us assume that $P(\vb{t}|\vb{x})$ has a unimodal distribution when $\vb{x}$ is fixed. If we approximate this distribution using Gaussian distribution, we can define $\Phi(\vb{x},\vb{t})$ as 
\begin{equation}
\label{eq: phi (Regression problem)}
    \Phi(\vb{x},\vb{t}) = \frac{1}{\sqrt{(2\pi)^n |\tilde{\vb{\Sigma}}|}} \exp\left[-\frac{1}{2}(\vb{t}-\boldsymbol{\mu})^\mathrm{T}\tilde{\vb{\Sigma}}^{-1}(\vb{t}-\boldsymbol{\mu})\right] \, ,
\end{equation}
where $n$ is the dimension of $\vb{t}$, $\tilde{\vb{\Sigma}}$ is a $n\times n$ matrix and $\boldsymbol{\mu}$ is a vector which has same dimension as $\vb{t}$. Here, $\tilde{\vb{\Sigma}}$ and $\boldsymbol{\mu}$ are the function of $\vb{x}$ and calculated through the model. Fig.\,\ref{fig: regression} shows an example case for $n=2$, where $\boldsymbol{\mu} = (\mu_1, \mu_2)$ and 
\begin{equation}
\label{eq: sigma (Regression problem)}
    \tilde{\boldsymbol{\Sigma}} = \left(
\begin{array}{cc}
\sigma_{11} & \sigma_{12}  \\
\sigma_{12}& \sigma_{22}\\
\end{array}
\right)\, .
\end{equation}
We can easily check that Eq.\,(\ref{eq: phi (Regression problem)}) satisfies the normalization condition in Eq.\,(\ref{eq: norm condition (Regression problem)}). Then the loss function is calculated as
\begin{equation}
\label{eq: expanded loss (Regression problem)}
   L(\Phi(\vb{x},\vb{t})) = \frac{n}{2}\log{2\pi} + \frac{1}{2}\log{|\tilde{\vb{\Sigma}}|} + \frac{1}{2}(\vb{t}-\boldsymbol{\mu})^\mathrm{T}\tilde{\vb{\Sigma}}^{-1}(\vb{t}-\boldsymbol{\mu}) \, ,
\end{equation}
where the first term can be neglected because it is just a constant. This is the loss function we should use, when we assume that probability distribution $P(\vb{t}|\vb{x})$ can be approximated by a Gaussian distribution. If we consider further approximation of fixing $\tilde{\vb{\Sigma}}$ to be an identity matrix, the loss function is simplified as
\begin{equation}
\label{eq: mean squared error loss (Regression problem)}
    L(\Phi(\vb{x},\vb{t})) = \frac{n}{2}\log{2\pi} + \frac{1}{2}(\vb{t}-\boldsymbol{\mu})^2 \, ,
\end{equation}
and this is equivalent to mean squared error. The true meaning of using mean squared error is approximating $P(\vb{t}|\vb{x})$ by a Gaussian distribution with constant variances. 

So far we have been discussing about the approximation by Gaussian distribution, but this is just an example. We should use other distributions such as Cauchy distribution or gamma distribution, depending on the property of the data set.     

\subsection{Parameter estimation problem}
\label{sec: Parameter estimation problem}
 Suppose that we are given some data sets which are denoted as $\{\vb{x}_1,\cdots,\vb{x}_N\}$.
 If these data sets are considered to follow specific distribution parametrized by some variables, how to estimate these variables? To solve such a problem, maximum likelihood estimation is often used, and a brief overview will be given. First, one assume certain probability distribution with some unknown parameters. For example, if data set is considered to obey a Gaussian distribution, the mean value and variance of the distribution would be good parameters. Let $\boldsymbol{\theta}$ be the set of the parameters, and $P(\vb{x}|\boldsymbol{\theta})$ be the probability distribution of $\vb{x}$ parametrized by $\boldsymbol{\theta}$. Our goal is to find the best $\boldsymbol{\theta}$ which explain the data set. This is done by maximizing the conditional probability of getting the data set $\{\vb{x}_1,\cdots,\vb{x}_N\}$ under the assumption of certain $\boldsymbol{\theta}$. It means maximizing the likelihood $L^*$ by optimizing $\boldsymbol{\theta}$, and it is written as 
\begin{equation}
\label{eq: likelihood (Parameter estimation problem)}
    L^*(\boldsymbol{\theta}) = \prod_{i=1}^{N}P(\vb{x}_i|\boldsymbol{\theta}) \, .
\end{equation}
It is also fine to maximize the logarithm of the likelihood written as
\begin{equation}
\label{eq: log likelihood (Parameter estimation problem)}
    \log{L^*(\boldsymbol{\theta})} = \sum_{i=1}^{N}\log{P(\vb{x}_i|\boldsymbol{\theta})} \, .
\end{equation}

\begin{figure}[t]
    \centering
    \includegraphics[scale=0.8]{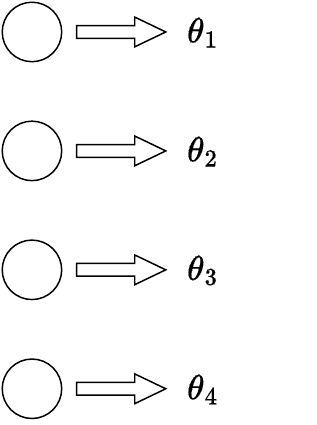}
    \caption{\small \sl The example of the model for parameter estimation problem. In this figure, we assume 4 parameters.}
    \label{fig: likelihood}
\end{figure}

Next, we will show that this problem can be solved based on learning principle. In this problem, the original purpose is to find the true probability distribution $P(\vb{x})$ that the data set follows. In other words, this problem is a type of the problem of finding probabilities by unsupervised learning, as discussed in Sec.\,\ref{sec: General case}. Using the conclusions there, we consider the model function $\Phi(\vb{x})$, and define the loss function as
\begin{equation}
\label{eq: log loss (Parameter estimation problem)}
    L(\Phi(\vb{x})) = - \log{(\Phi(\vb{x}))}\, ,
\end{equation}
under the normalization condition as
\begin{equation}
\label{eq: norm condition (Parameter estimation problem)}
    \int d\vb{x} \Phi(\vb{x})  = 1 \, .
\end{equation}
To satisfy this condition, we use the same assumptions as in the maximum likelihood estimation, and we can write it as
\begin{equation}
\label{eq: phi (Parameter estimation problem)}
    \Phi(\vb{x}) = P(\vb{x}|\boldsymbol{\theta})\, .
\end{equation}
This can be interpreted as the model is $\boldsymbol{\theta}$ itself and no other parameters. Fig.\,\ref{fig: likelihood} shows the model for this problem, though there are no layers or connections. Of course, this equation satisfies the normalization condition in Eq.\,(\ref{eq: norm condition (Parameter estimation problem)}).
Then the loss function is written as
\begin{equation}
\label{eq: expanded loss (Parameter estimation problem)}
    L(\Phi(\vb{x})) = -\log P(\vb{x}|\boldsymbol{\theta}) \, .
\end{equation}

Comparing Eq.\,(\ref{eq: log likelihood (Parameter estimation problem)}) and Eq.\,(\ref{eq: expanded loss (Parameter estimation problem)}), the former takes the sum over all the data in the data set, while the latter only performs the calculation on each piece of data. The difference in sign is whether you want to maximize or minimize the value. Here, our argument is that we only need to optimize the parameter $\boldsymbol{\theta}$ for each data one by one, and there is no need to compute sum over the entire data set together. Both methods will give same $\boldsymbol{\theta}$ finally, but our method is faster and more flexible because it does not need to take summation. 

\section{Normalization by differential}
\label{sec: Normalization by differential}
As we have seen, the loss function is always written as $L(\Phi(\vb{x})) = -\log(\Phi(\vb{x}))$, and the essence of the problem is how to make $\Phi(\vb{x})$ satisfy the normalization condition. If this normalization condition is satisfied, the learning principle is easily achieved. In cases where the normalization condition includes integration, usually some assumptions and approximations are used. In this chapter, we show how to make $\Phi(\vb{x})$ satisfy the normalization condition without any assumptions or approximations. Using the method discussed in this chapter, it is possible to learn any probability distribution without any prior knowledge, and this is a universal learning method in the true sense.

\subsection{Input with a single variable}
\label{sec: Input with a single variable}

\begin{figure}[t]
    \centering
    \includegraphics[scale=0.8]{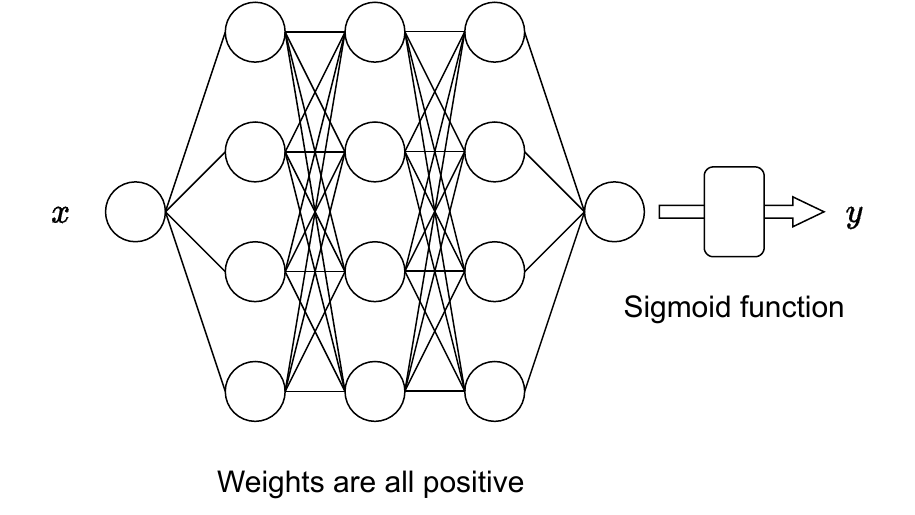}
    \caption{\small \sl The example of the model to use the differentiation method. This model is for the input with one variable.}
    \label{fig: differential1}
\end{figure}

First, for simplicity, we consider the case where each input has only one variable. That is, given a set of numbers $\{x_1,x_2,\cdots\}$, and we estimate the probability distribution $P(x)$ they follow.
In this case, our goal is to construct the model function $\Phi(x)$ which satisfies the normalization condition as
\begin{equation}
\label{eq: norm condition (Input with a single variable)}
    \int dx \, \Phi(x) = 1 \, ,
\end{equation}
without any assumptions or approximations. The idea for this is very simple. We need a normalization condition for integrated value, so we take the differential in advance to define the model function. Specifically, considering the model shown in the Fig.\,\ref{fig: differential1}, and we define $\Phi(x)$ as
\begin{equation}
\label{eq: derivative one parameter (Input with a single variable)}
    \Phi(x) = \frac{dy(x)}{dx} \, ,
\end{equation}
where $y$ is the value of the final output. Since it should denote the estimation of probability, we impose additional conditions as 
\begin{gather}
\label{eq: monotonic one parameter (Input with a single variable)}
    \frac{dy}{dx}\geq 0 \, , \\
\label{eq: range (Input with a single variable)}
    0 \leq \,y(x)\leq 1 \, .  
\end{gather}

The condition in Eq.\,(\ref{eq: monotonic one parameter (Input with a single variable)}) is introduced to make $\Phi(x)$ positive, and it can be achieved by using only positive weights and activation functions which is monotonically increasing. The condition in Eq.\,(\ref{eq: range (Input with a single variable)}) can be achieved by using Sigmoid function at the final node. Then we can check the normalization condition as 
\begin{equation}
\label{eq: inequality (Input with a single variable)}
\begin{split}
    \int dx \, \Phi(x) &= \int dx \, \frac{d y(x)}{dx} \\
    &= \int_0^1 dy \\ 
    &= 1 \, .
\end{split}
\end{equation}
Of course, the loss function is defined as
\begin{equation}
\label{eq: log loss (Input with a single variable)}
    L(\Phi(x)) = -\log{(\Phi(x))} \, .    
\end{equation}
As long as $\Phi(x)$ has a universal approximation property, $\Phi(x)$ approaches the true probability distribution $P(x)$ as the learning proceeds. No assumptions or approximations are used here, and no prior knowledge of the data is required. As a note, the value obtained by differentiating $y(x)$ with respect to $x$ corresponds to the probability distribution function, so $y(x)$ corresponds to the cumulative distribution function.

\begin{figure}[H]
  \begin{minipage}[t]{0.48\hsize}
    \centering
    \includegraphics[width = 86mm]{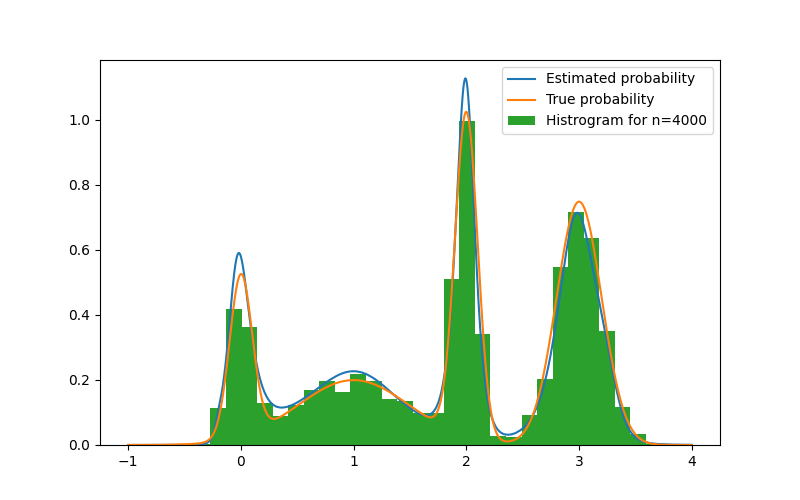}
  \end{minipage}
  \begin{minipage}[t]{0.48\hsize}
    \centering
    \includegraphics[width = 86mm]{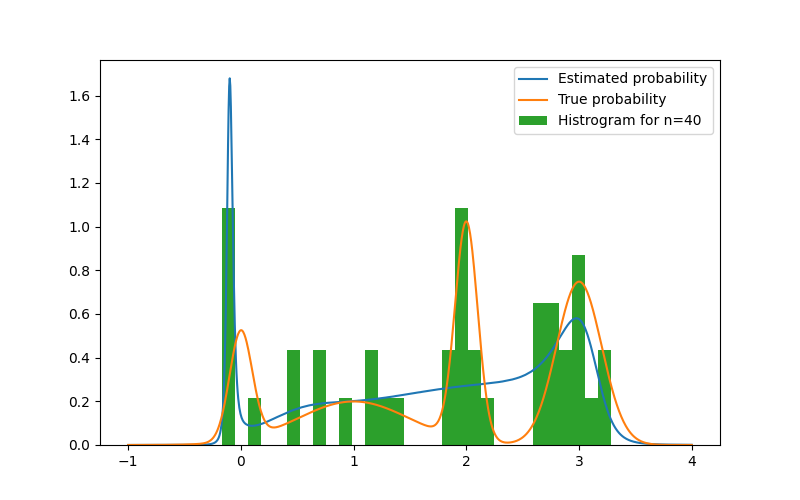}
  \end{minipage}
  \caption{\small \sl Probability estimation for multimodal distribution. The sample size for left figure is 4000 and for right figure is 40. True probabilities are same for both figures.}
\label{fig: demonstration1}
\end{figure}

\begin{figure}[H]
  \begin{minipage}[t]{0.48\hsize}
    \centering
    \includegraphics[width = 86mm]{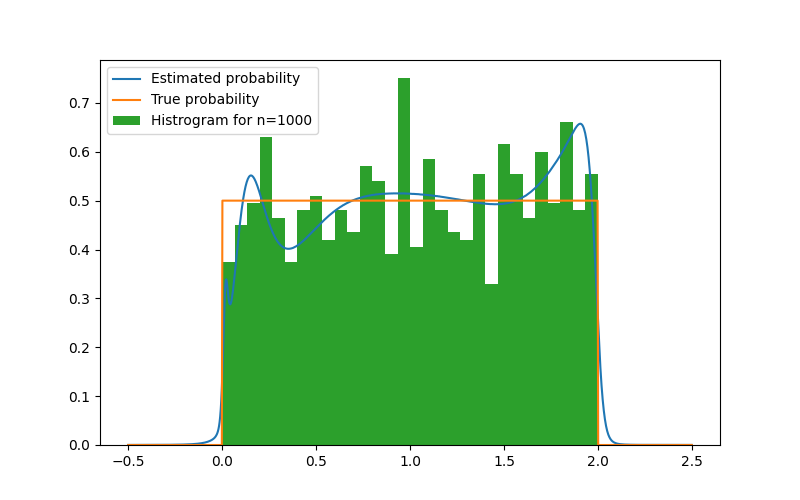}
  \end{minipage}
  \begin{minipage}[t]{0.48\hsize}
    \centering
    \includegraphics[width = 86mm]{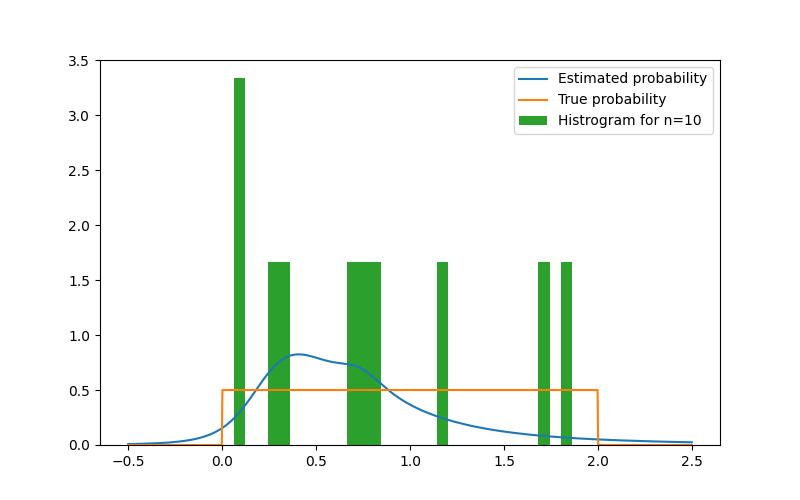}
  \end{minipage}
  \caption{\small \sl Probability estimation for flat distribution. The sample size for left figure is 1000 and for right figure is 10. True probabilities are same for both figures.}
\label{fig: demonstration2}
\end{figure}

\begin{figure}[H]
  \begin{minipage}[t]{0.48\hsize}
    \centering
    \includegraphics[width = 86mm]{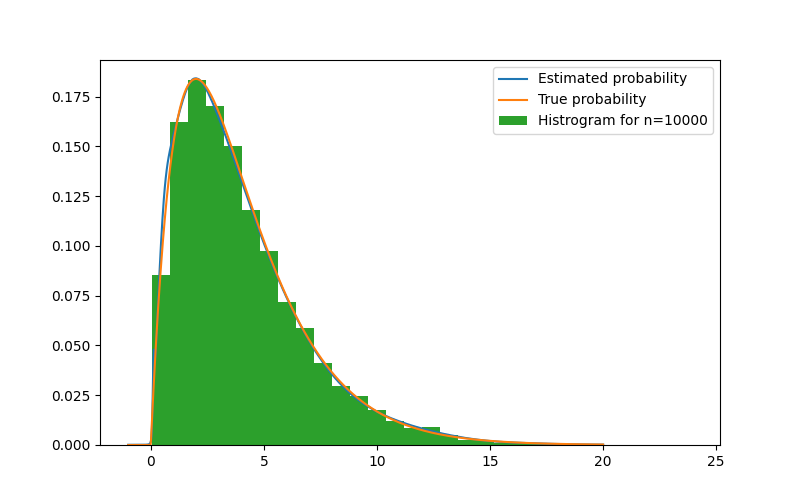}
  \end{minipage}
  \begin{minipage}[t]{0.48\hsize}
    \centering
    \includegraphics[width = 86mm]{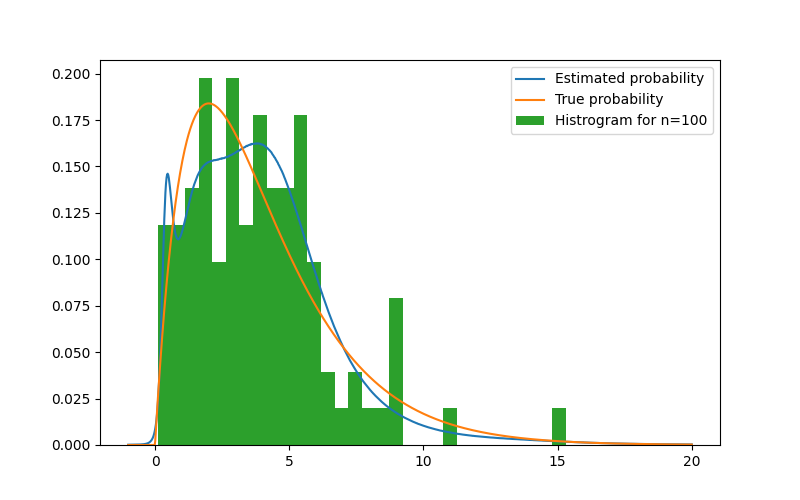}
  \end{minipage}
  \caption{\small \sl Probability estimation for skewed distribution. The sample size for left figure is 10000 and for right figure is 100. True probabilities are same for both figures.}
\label{fig: demonstration3}
\end{figure}

To demonstrate how powerful this new method is, the results of the calculations on several data sets are shown in Figs.\,\ref{fig: demonstration1},\ref{fig: demonstration2},\ref{fig: demonstration3}. In these figures, random samples are selected from the probability distribution drawn with orange lines, and the samples are shown by green histogram. We estimate the probability distribution from these samples using our method, and it is drawn with blue lines. As you can see, the fittings are very good for any distribution, even if it is multimodal or sparse. 

\subsection{Input with multiple variables}
\label{sec: Input with multiple variables}
\begin{figure}[t]
    \centering
    \includegraphics[scale=0.8]{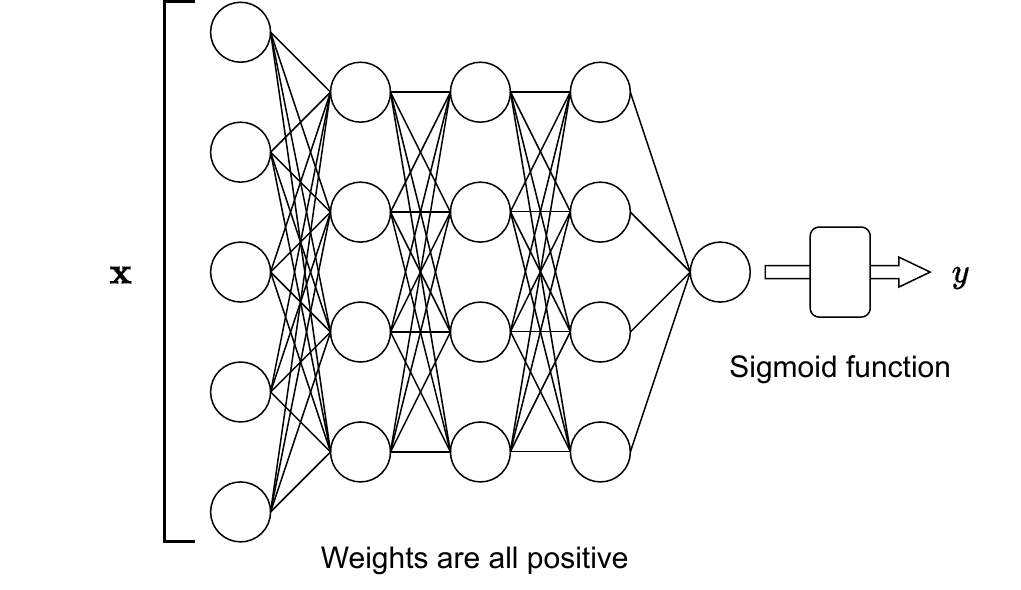}
    \caption{\small \sl  The example of the model to use the differentiation method. This model is for the input with multiple variables.}
    \label{fig: differential2}
\end{figure}

Next, we consider the case where each input contains multiple variables. We take 
$\{\vb{x}_1,\vb{x}_2,\cdots,\}$ as input data set, where $\vb{x} = (a_1,a_2,\cdots, a_n)$ and each $a_i$ takes continuous value. To find $P(\vb{x})$ without assumptions or approximations, let us consider a generalization of the method derived in the previous section for multiple variables.
Considering the model defined as shown in Fig.\,\ref{fig: differential2}, and we define the model function $\Phi(\vb{x})$ as
\begin{equation}
\label{eq: phi (Input with multiple variables)}
    \Phi(\vb{x}) = \frac{\partial^n y(\vb{x})}{\partial a_1 \partial a_2 \cdots \partial a_n} \, ,
\end{equation}
where $y(\vb{x})$ is the value at the final layer.
As same as the previous section, we impose the condition to $y(\vb{x})$ as
\begin{gather}
\label{eq: monotonic (Input with multiple variables)}
    \frac{\partial^n y(\vb{x})}{\partial a_1 \partial a_2 \cdots \partial a_n} \geq 0 \, ,\\
\label{eq: range (Input with multiple variables)}
    0 \leq \, y(\vb{x}) \leq 1 \, .
\end{gather}
These conditions are achieved by positive weights and sigmoid function as same as previous section.
We can easily check the normalization condition as
\begin{equation}
\label{eq: norm condition (Input with multiple variables)}
\begin{split}
    \int d\vb{x} \,\Phi(\vb{x}) &=  \int \prod_{i=1}^{n}da_i \, \frac{\partial^n y(\vb{x})}{\partial a_1 \partial a_2 \cdots \partial a_n} \\
    &= \int_0^1 dy \\ 
    &= 1 \, .
\end{split}
\end{equation}
The loss function is defined as 
\begin{equation}
\label{eq: log loss (Input with multiple variables)}
    L(\Phi(\vb{x})) = -\log{(\Phi(\vb{x}))} \, .
\end{equation}
This model looks good, but is impractical because Eq.\,(\ref{eq: phi (Input with multiple variables)}) contains $n$-th derivative of $y(\vb{x})$. If $n$ becomes large, the numerical calculation of the multiple differentiation is impossible. In the following section, we suggest the solution for this problem.

\subsection{Input with multiple variables and output with multiple variables}
\label{sec: Input with multiple variables and output with multiple variables}

\begin{figure}[t]
    \centering
    \includegraphics[scale=0.8]{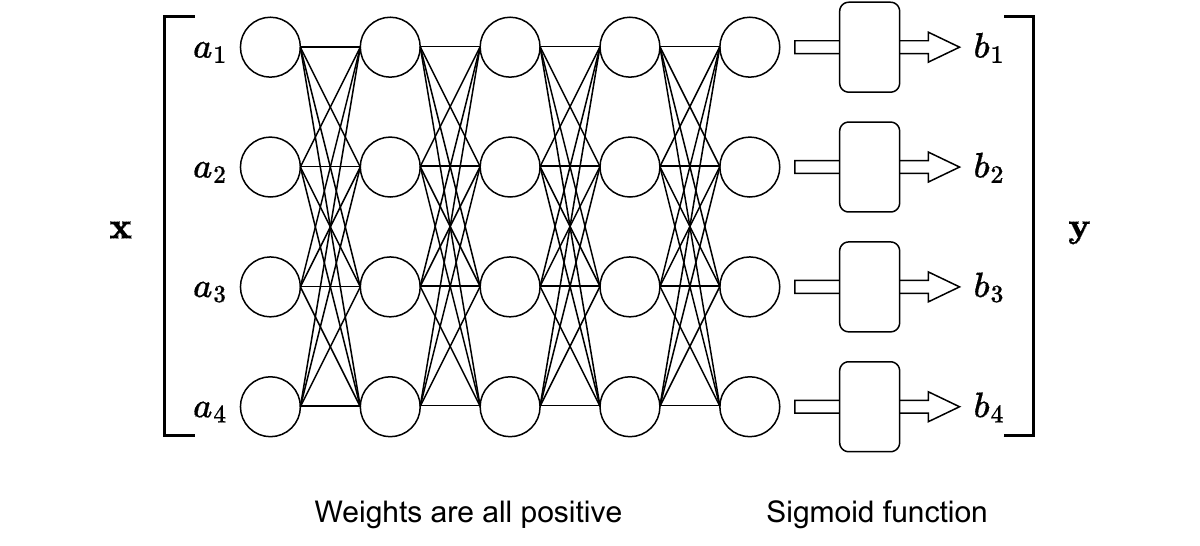}
    \caption{\small \sl The example of the model to use the differentiation method. This model is for the input with multiple variables and output with multiple variables.}
    \label{fig: differential3}
\end{figure}
\begin{figure}[t]
    \centering
    \includegraphics[scale=0.8]{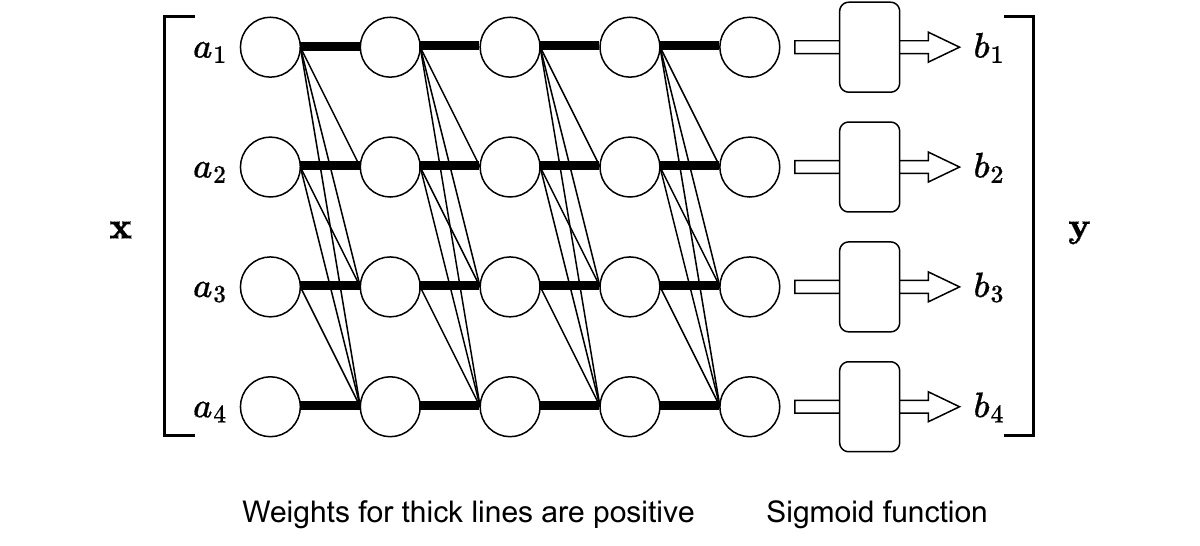}
    \caption{\small \sl The example of the model to use the differentiation method. This model is for the input with multiple variables and output with multiple variables. In this model, there is no edge from lower nodes to upper nodes, and horizontal edges have a positive weight.}
    \label{fig: differential4}
\end{figure}

 
The problem in the previous section was caused by the inclusion of multiple derivatives in the definition of $\Phi(\vb{x})$. To solve this, we propose to use Jacobian determinant as another generalization of Eq.\,(\ref{eq: derivative one parameter (Input with a single variable)}) to multiple variables. For this purpose, we consider the model shown in Fig.\,\ref{fig: differential3} and define the model function $\Phi(\vb{x})$ as
\begin{equation}
\label{eq: phi (Input with multiple variables and output with multiple variables)}
    \Phi(\vb{x}) = \frac{\partial\vb{y}(\vb{x})}{\partial\vb{x}}\, ,
\end{equation}
where $\vb{y}(\vb{x}) = (b_1, b_2, \cdots, b_n)$ denotes the variables in the final layer and
right hand side denotes Jacobian determinant. This determinant is defined as
\begin{equation}
\label{eq: jacobian (Input with multiple variables and output with multiple variables)}
    \frac{\partial\vb{y}(\vb{x})}{\partial\vb{x}} \equiv 
\left|
\begin{array}{ccc}
\displaystyle \frac{\partial b_1}{\partial a_1} & \displaystyle \cdots & \displaystyle \frac{\partial b_1}{\partial a_n} \\
\displaystyle \vdots& \displaystyle \ddots & \displaystyle \vdots\\
\displaystyle  \frac{\partial b_n}{\partial a_1} & \displaystyle \cdots & \displaystyle\frac{\partial b_n}{\partial a_n} \\
\end{array}
\right|\, .
\end{equation}
If we impose the condition to $\vb{y}(\vb{x})$ as
\begin{equation}
\label{eq: monotonic (Input with multiple variables and output with multiple variables)}
\begin{split}
    0\leq b_1,&b_2,\cdots,b_n \leq 1 \, ,
\end{split}    
\end{equation}
then we can check the normalization condition of $\Phi(\vb{x})$ as
\begin{equation}
\label{eq: norm condition (Input with multiple variables and output with multiple variables)}
\begin{split}
\int d\vb{x} \,\Phi(\vb{x}) &= \int \prod_{i=1}^n da_i\frac{\partial\vb{y}(\vb{x})}{\partial\vb{x}} \\
&= \int_{0}^{1} \prod_{i=1}^n db_i \\
&= 1 \, .
\end{split}
\end{equation}
This $\Phi(\vb{x})$ only contains first derivatives, and the problem mentioned in Sec.\,\ref{sec: Input with multiple variables} seems to be solved, but other problems occur. First, Eq.\,(\ref{eq: jacobian (Input with multiple variables and output with multiple variables)}) is the determinant of $n$-dimensional matrix, and the order of its calculation is $\mathcal{O}(n^3)$, then it is still impractical for large $n$. Second, there is no guarantee that this Jacobian determinant is positive and $\log(\Phi(\vb{x}))$ cannot be calculated if $\Phi(\vb{x})$ is negative.
To solve these problems, we impose new conditions as  
\begin{gather}
\label{eq: zero (Input with multiple variables and output with multiple variables)}
    i<j \Rightarrow \frac{\partial b_i}{\partial a_j} = 0  \, , \\
\label{eq: diagonal (Input with multiple variables and output with multiple variables)}
  \forall i:  \frac{\partial b_i}{\partial a_i} > 0  \, .  
\end{gather}
These conditions are achieved using the model defined in Fig.\,\ref{fig: differential4}.
Then the Jacobian matrix becomes triangular matrix and  $\Phi(\vb{x})$ is calculated as 
\begin{equation}
\label{eq: triangular matrix (Input with multiple variables and output with multiple variables)}
\begin{split}
    \Phi(\vb{x}) &=  
\left|
\begin{array}{ccc}
\displaystyle \frac{\partial b_1}{\partial a_1} & \displaystyle \cdots &  \displaystyle \text{\Large{0}} \\
\displaystyle \vdots& \displaystyle \ddots & \displaystyle \vdots\\
\displaystyle  \frac{\partial b_n}{\partial a_1} & \displaystyle \cdots & \displaystyle\frac{\partial b_n}{\partial a_n} \\
\end{array}
\right| \\
& = \prod_{i=1}^{n} \frac{\partial b_i}{\partial a_i} \, .
\end{split}    
\end{equation}
$\Phi(\vb{x})$ becomes just a product of diagonal elements, so the order of calculation is $\mathcal{O}(n)$ and it is also guaranteed to be positive because of Eq.\,(\ref{eq: diagonal (Input with multiple variables and output with multiple variables)}). In this case, the loss function is defined as
\begin{equation}
\label{eq: expanded loss (Input with multiple variables and output with multiple variables)}
\begin{split}
    L(\Phi(\vb{x}))&=-\log(\prod_{i=1}^{n} \frac{\partial b_i}{\partial a_i} ) \\
    &= -\sum_{i=1}^{n}\log(\frac{\partial b_i}{\partial a_i}) \, .
\end{split}
\end{equation}

Next, we discuss the meaning of each diagonal element ${\partial b_i}/{\partial a_i}$. First, $b_1$ only depends on $a_1$ and is independent of $(a_2,\cdots,a_n)$. This situation is completely same as one parameter case discussed in Sec.\,\ref{sec: Input with a single variable}. Then we can conclude 
\begin{equation}
\label{eq: a1 b1 (Input with multiple variables and output with multiple variables)}
    \frac{\partial b_1}{\partial a_1} \to P(a_1)\, ,
\end{equation}
where $P(a_1)$ is the probability of the emergence of $a_1$. Second, $b_2$ only depends on $(a_1, a_2)$ and is independent of $(a_3,\cdots,a_n)$. Then we can conclude
\begin{equation}
\label{eq: a1 b1 a2 b2 (Input with multiple variables and output with multiple variables)}
    \frac{\partial b_1}{\partial a_1} \frac{\partial b_2}{\partial a_2}\to P(a_1,a_2) \, ,
\end{equation}
where $P(a_1, a_2)$ is the probability of the emergence of $(a_1,a_2)$. From these two formulae, we get the relation as
\begin{equation}
\label{eq: a2 b2 (Input with multiple variables and output with multiple variables)}
   \frac{\partial b_2}{\partial a_2}\to \frac{P(a_1,a_2)}{P(a_1)} = P(a_2|a_1) \, .
\end{equation} 
 We can repeat this 
calculation, and we conclude that
\begin{equation}
\label{eq: ai bi (Input with multiple variables and output with multiple variables)}
    \frac{\partial b_i}{\partial a_i} \to P(a_i|(a_1,\cdots,a_{i-1})) \, .
\end{equation}

\subsection{Regression problem for general case}
\label{sec: Regression problem for general case}
\begin{figure}[t]
    \centering
    \includegraphics[scale=0.8]{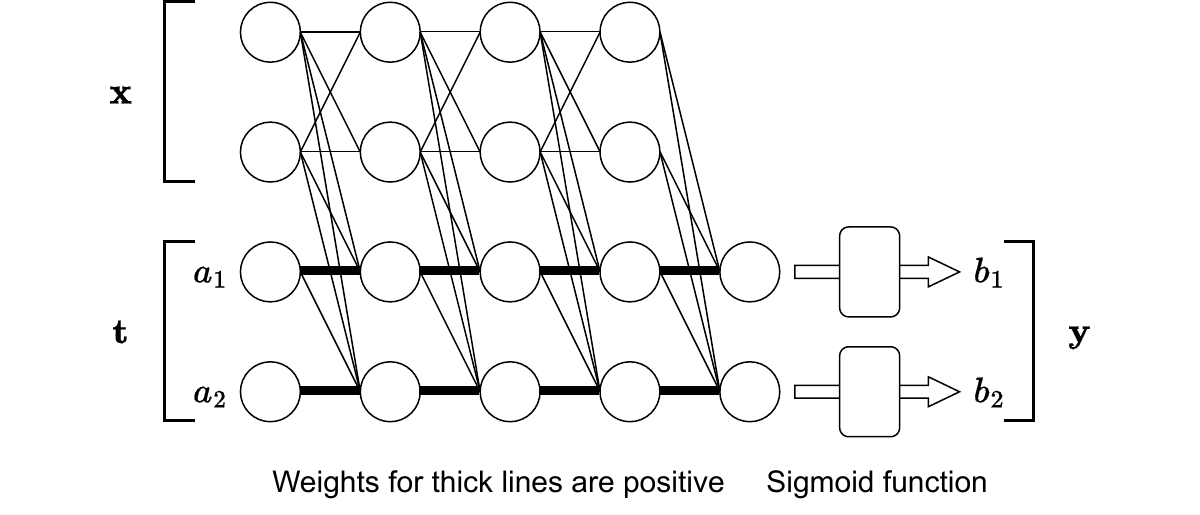}
    \caption{\small \sl The example of the model for regression problem using differential method.}
    \label{fig: differential5}
\end{figure} 
This section discusses how to solve regression problems discussed in Sec.\,\ref{sec: Regression problem} without assumptions or approximations. The goal is to construct $\Phi(\vb{x},\vb{t})$ such that Eq.\,(\ref{eq: norm condition (Regression problem)}) is always satisfied. With reference to the method proposed in the previous section, we consider the model shown in the Fig.\,\ref{fig: differential5}, and define the model function $\Phi(\vb{x},\vb{t})$ as
\begin{equation}
\label{eq: phi (Regression problem for general case)}
    \Phi(\vb{x},\vb{t}) = \frac{\partial\vb{y}(\vb{x},\vb{t})}{\partial \vb{t}} \, ,
\end{equation}
 where we suppose $\vb{t}$ has $n$ variables as $\vb{t} = (a_1, a_2,\cdots,a_n)$ and $\vb{y}(\vb{x},\vb{t})$ also has $n$ variables as  $\vb{y}(\vb{x},\vb{t}) = (b_1,b_2,\cdots,b_n)$.
In addition, we impose the following conditions such as
\begin{gather}
\label{eq: zero (Regression problem for general case)}
    i<j \Rightarrow \frac{\partial b_i}{\partial a_j} = 0  \, , \\
\label{eq: diagonal (Regression problem for general case)}
     \forall i:  \frac{\partial b_i}{\partial a_i} > 0 \, ,  
\end{gather}
then $\Phi(\vb{x},\vb{t})$ is written as
\begin{equation}
\label{eq: triangular matrix (Regression problem for general case)}
\begin{split}
    \Phi(\vb{x},\vb{t}) &=  
\left|
\begin{array}{ccc}
\displaystyle \frac{\partial b_1}{\partial a_1} & \displaystyle \cdots &  \displaystyle \text{\Large{0}} \\
\displaystyle \vdots& \displaystyle \ddots & \displaystyle \vdots\\
\displaystyle  \frac{\partial b_n}{\partial a_1} & \displaystyle \cdots & \displaystyle\frac{\partial b_n}{\partial a_n} \\
\end{array}
\right| \\
& = \prod_{i=1}^{n} \frac{\partial b_i}{\partial a_i} \, .
\end{split}    
\end{equation}
The normalization condition is confirmed as
\begin{equation}
\label{eq: norm condition (Regression problem for general case)}
    \begin{split}
        \int d\vb{t} \Phi(\vb{x},\vb{t}) &= \int \prod_{i=1}^n da_i \frac{\partial b_i}{\partial a_i} \\
        &= \int_0^1 \prod_{i=1}^n db_i \\
        &=1 \, .
    \end{split}
\end{equation}
The loss function is defined as
\begin{equation}
\label{eq: expanded loss (Regression problem for general case)}
\begin{split}
    L(\Phi(\vb{x}))&=-\log(\prod_{i=1}^{n} \frac{\partial b_i}{\partial a_i} ) \\
    &= -\sum_{i=1}^{n}\log(\frac{\partial b_i}{\partial a_i}) \, .
\end{split}
\end{equation}

\subsection{Consideration about the normalization by differential}
\label{sec: Remarkable property of the normalization by differential}
 In this section, we note the remarkable property of the method proposed in this chapter. Since we want to discuss the most general case, we focus the model and loss function constructed in Sec.\,\ref{sec: Input with multiple variables and output with multiple variables}. We name the middle layers of the model in Fig.\,\ref{fig: differential4} as $(\vb{z}_1,\cdots,\vb{z}_d)$ where $d$ denotes the depth of the layers. Each layer has $n$ variables defined as $\vb{z}_i = (c_{1;i},c_{2;i},\cdots,c_{n;i})$. Using the chain rule of Jacobian determinant, $\Phi(\vb{x})$ is calculated as
\begin{equation}
\label{eq: phi (Remarkable property of the normalization by differential)}
\begin{split}
    \Phi(\vb{x}) &= \frac{\partial\vb{y}(\vb{x})}{\partial{\vb{x}}} \\
    &= \frac{\partial\vb{z}_1(\vb{x})}{\partial{\vb{x}}} \frac{\partial\vb{z}_2(\vb{z}_1)}{\partial{\vb{z}_1}} \cdots \frac{\partial\vb{z}_d(\vb{z}_{d-1})}{\partial{\vb{z}_{d-1}}}\frac{\partial\vb{y}(\vb{z}_d)}{\partial{\vb{z}_d}} \\
    & = \prod_{i=0}^{d} \prod_{j=1}^{n} \frac{\partial c_{j;i+1}}{\partial c_{j;i}} \, ,
\end{split}
\end{equation}
where we set $c_{j;0} = a_j$ and $c_{j;d+1} = b_j$ in the final line. Then, the loss function is calculated as 
\begin{equation}
\label{eq: expanded loss (Remarkable property of the normalization by differential)}
\begin{split}
    L(\Phi(\vb{x})) &= -\log\left[ \prod_{i=0}^{d} \prod_{j=1}^{n} \frac{\partial c_{i;j+1}}{\partial c_{i;j}}\right] \\
    &= - \sum_{i=0}^{d} \sum_{j=1}^{n} \log\frac{\partial c_{i;j+1}}{\partial c_{i;j}} \, .
\end{split}
\end{equation}
This equation suggest that the loss function is represented as the summation of localized loss function on each node. We can write localized loss function for each node $\mathcal{L}_{i,j}$ as
\begin{equation}
\label{eq: localized loss (Remarkable property of the normalization by differential)}
    \mathcal{L}_{i,j} = - \log\frac{\partial c_{i;j+1}}{\partial c_{i;j}} \, .
\end{equation}
In this case, there is no need to use back propagation\,\cite{rumelhart1986learning} to optimize the internal parameters. This is because it is sufficient to locally optimize each neighboring parameter related to the localized loss function. This can be understood as an analogy with dynamical systems in physics. For example, suppose you pick up both ends of a string that has mass and lift it under gravity. It is known that a string takes the form of a catenary line, and  this is when the potential energy of the string is at a minimum. It is not that some control tower is giving instructions to take this shape, but that the overall energy is naturally minimized as each local part of the string tries to minimize its energy. Exactly the same thing is achieved in this model as well. Furthermore, looking at the localized loss function in this model, there is no longer a concept of an output. That is to say, we can add as many layers as we want without changing the form of the localized loss function. These characteristics are very important in understanding the workings of the human brain, which will be discussed in the next chapter.

\section{Normalization by time evolution}
\label{sec: Normalization by time evolution}
In this chapter, we will focus on constructing a model that mimics the learning mechanism in the brain. As we have discussed in previous chapters, learning is all about estimating the probability of the input, and the human brain is no exception. If we consider the brain as a kind of machine learning model, there is a certain model function $\Phi(\vb{x})$ for a given input $\vb{x}$, where $\Phi(\vb{x})$ satisfies the normalization condition as
\begin{equation}
\label{eq: norm condition (Normalization by time evolution)}
    \int d\vb{x} \, \Phi(\vb{x}) = 1 \, .
\end{equation}
The question is how this normalization condition is implemented in the brain, and this will be the focus of the discussion. 

\begin{figure}[t]
    \centering
    \includegraphics[scale=0.8]{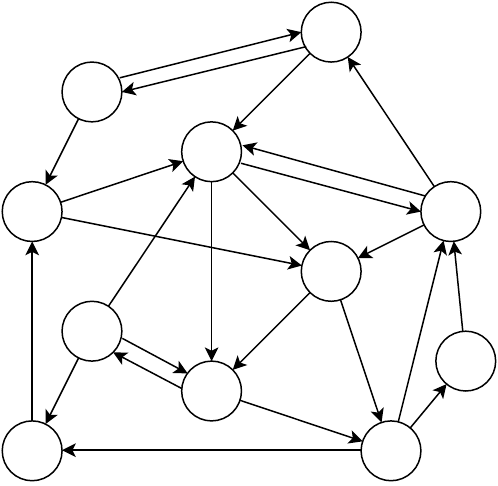}
    \caption{\small \sl The example of a fully or partially connected model.}
    \label{fig: allconnected}
\end{figure}

There are several conditions that must be met when creating a model that mimics the human brain. First, the model must be general-purpose. The human brain processes information from the five senses and automatically learns the rules behind it. In other words, it is performing unsupervised learning without prior knowledge. Second, nodes and edges must be homogeneous and isotropic, with no special nodes or special edges. This is due to the observation that neurons in the brain appear to form fully or partially connected networks. Neurons in the brain do not necessarily form layers and may have a loop structure. In that sense, the model discussed in Chapter\,\ref{sec: Normalization by differential}, although general-purpose, cannot be a model of the human brain. Based on this request, we consider a fully or partially connected model as shown in Fig.\ref{fig: allconnected}. 
As an analogy for synapses in the brain having orientation, we assume that each edge has a orientation, and that information is transmitted only in this direction. Of course, edges may be connected in any way, and there may be no edges between certain nodes, or there may be edges in both directions. As will be shown later, by considering the time evolution of such a model, it is possible to construct $\Phi(\vb{x})$ that satisfies the normalization conditions. As time progresses, each node changes its internal state as it is influenced by other nodes, and this flow itself is considered as a machine learning model. In the next section, we will concretely construct $\Phi(\vb{x})$ and confirm that the normalization condition is satisfied.

\subsection{Linear model}
\label{sec: Linear model}
Before going into detailed calculations, first we define the variables in the model such as in Fig.\,\ref{fig: allconnected}. We suppose there are $n$ nodes in the model, and they change in a time-dependent manner. We name them as $\vb{a}(t) = (a_1(t),\cdots,a_n(t))$, where $t$ is a variable representing time and each $a_i(t)$ takes a real number. Next, each edge has a weight represented by a real number, and let $w_{ij}$ be the weight of the edge from the $j$-th to the $i$-th node. Here, no self-coupling of nodes is assumed, so $w_{ii} = 0$ accordingly. When we display $w_{ij}$ together as a matrix, we will use the symbol $\tilde{\vb{W}}$. Finally, each node has a constant bias value, which is defined as $\vb{b} = (b_1,\cdots, b_n)$. Only $\vb{a}(t)$ changes in a time-dependent manner among these variables. $\tilde{\vb{W}}$ and $\vb{b}$ are the targets of optimization. 

Next, we define the time evolution of this model. First, input $\vb{x}$ is entered into the model at $t=0$ and time evolution is started from there. Since there is no special input layer in this model, some of the nodes must be assigned as input. Let us assume that $\vb{x}$ consists of $m$ variables written as $\vb{x} = (x_1,\cdots,x_m)$, and we assign $(a_1(0),\cdots,a_m(0)) = (x_1,\cdots,x_m)$. For the remaining nodes, we assign $(a_{m+1}(0),\cdots,a_n(0)) = (r_{m+1},\cdots,r_n) $, where $r_i$ is a random value which obeys certain probability distribution
and is chosen randomly every time. 

\begin{figure}[t]
    \centering
    \includegraphics[scale=0.8]{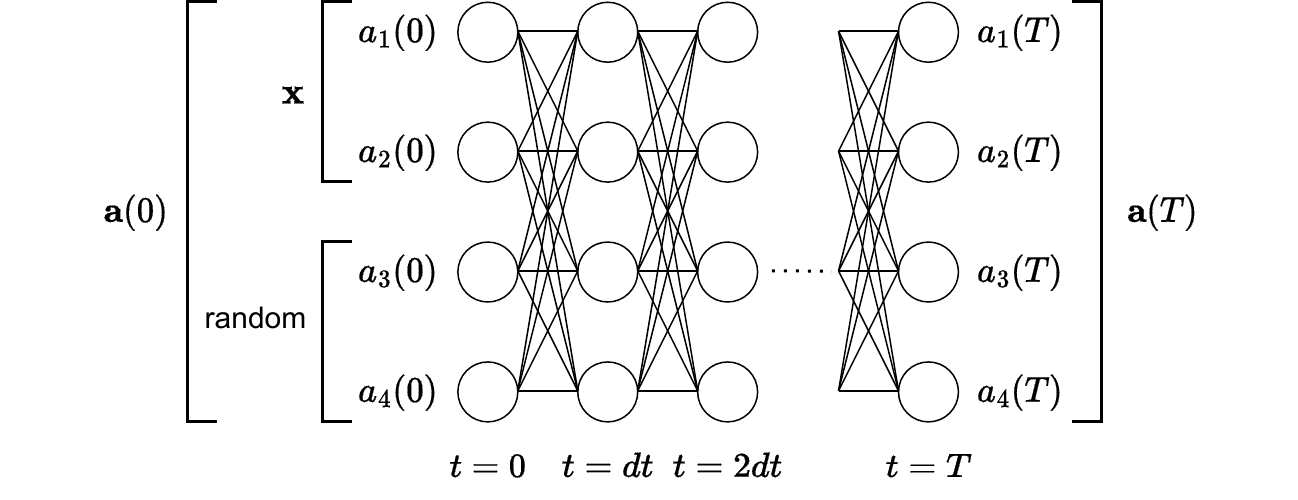}
    \caption{\small \sl The conceptual diagram which illustrate the time evolution of the fully or partially connected model. In this figure, we assume $n$=4 and $m$=2.}
    \label{fig: allconnected_concept}
\end{figure}

Now that the initial conditions have been established, we need the equations which describe the time evolution. They are defined as
\begin{equation}
\label{eq: time evolution (Linear model)}
    a_i(t+dt) - a_i(t) =   (\sum_{j=1}^n w_{ij} a_j(t)  + b_i) dt \, .
\end{equation}
This time evolution equation also can be written in vector form as
\begin{equation}
\label{eq: time evolution vec (Linear model)}
    \vb{a}(t+dt) - \vb{a}(t) = (\tilde{\vb{W}}\,\vb{a}(t) + \vb{b} )dt \,  .
\end{equation}
Then we let this model evolve over time until $t=T$, where $T$ is chosen arbitrary. The time evolution in this model plays the role of layers in ordinary machine learning, and Fig.\,\ref{fig: allconnected_concept} illustrates this concept.
Next, instead of $\Phi(\vb{x})$, we try to construct the model function $\Phi(\vb{a}(0))$ which satisfies the normalization condition as
\begin{equation}
\label{eq: norm condition 1 (Linear model)}
     \int^{\infty}_{-\infty} d\vb{a}(0) \,\Phi(\vb{a}(0)) = 1\, .
\end{equation}
If this condition is satisfied, $\Phi(\vb{a}(0))$ is expected to approach $P(\vb{a}(0))$ as the learning proceed, where $P(\vb{a}(0))$ is calculated as
\begin{equation}
\label{eq: probability (Linear model)}
    P(\vb{a}(0)) = P(\vb{x})\left(\prod_{i=m+1}^{n}P(a_i(0)=r_i)\right) \, .
\end{equation}
From this equation, since we already know the probability distribution for $r_i$, $P(\vb{x})$ can be calculated backwards. Now, let us define $\Phi(\vb{a}(0))$ as
\begin{equation}
\label{eq: phi (Linear model)}
    \Phi(\vb{a}(0)) = \frac{\exp[-E(\vb{a}(T))]}{\displaystyle\int^{\infty}_{-\infty} d\vb{a}\exp[-E(\vb{a})]} \, ,
\end{equation}
where $E(\vb{a})$ is an arbitrary function of $\vb{a}$ such that the integration in the denominator does not diverge. In the right hand side, $\vb{a}(T)$ is considered to be a function of $\vb{a}(0)$. 
Next, we prove that $\Phi(\vb{a}(0))$ satisfies the normalization condition. We first solve the differential equation in Eq.\,(\ref{eq: time evolution vec (Linear model)}), and we get
\begin{equation}
\label{eq: solution (Linear model)}
    \vb{a}(T) = \exp[\tilde{\vb{W}}T]\vb{a}(0)  + \left(\exp[\tilde{\vb{W}}T]-1 \right)\tilde{\vb{W}}^{-1} \vb{b}\, .
\end{equation}
Then the normalization condition is checked as
\begin{equation}
\label{eq: norm condition 2 (Linear model)}
\begin{split}
    \int^{\infty}_{-\infty} d\vb{a}(0) \,\Phi(\vb{a}(0)) 
    &=
    \left(\int_{-\infty}^{\infty} d\vb{a}(0)  \exp[-E(\vb{a}(T))]\right)  \left(\int_{-\infty}^{\infty} d\vb{a}  \exp[-E(\vb{a})]\right)^{-1} \\
    &=  \left(\int_{-\infty}^{\infty} d\vb{a}(T) \left(\frac{\partial \vb{a}(T)}{\partial\vb{a}(0)}\right)^{-1} \exp[-E(\vb{a}(T))]\right)  \left(\int_{-\infty}^{\infty} d\vb{a}  \exp[-E(\vb{a})]\right)^{-1} \\
    &= \left(\int_{-\infty}^{\infty} d\vb{a}(T) \left[\det(\exp[\tilde{\vb{W}}T])\right]^{-1} \exp[-E(\vb{a}(T))]\right)  \left(\int_{-\infty}^{\infty} d\vb{a}  \exp[-E(\vb{a})]\right)^{-1} \\
    &=\left[\det(\exp[\tilde{\vb{W}}T])\right]^{-1}\left(\int_{-\infty}^{\infty} d\vb{a}(T)  \exp[-E(\vb{a}(T))]\right)  \left(\int_{-\infty}^{\infty} d\vb{a}  \exp[-E(\vb{a})]\right)^{-1} \\
    &=\left[\exp(\tr[\tilde{\vb{W}}T])\right]^{-1} \\
    &= 1 \, ,
\end{split}
\end{equation}
where we use $\tr(\tilde{\vb{W}}T) = 0$ because $w_{ii} = 0$ in the final line. The loss function is calculated as 
\begin{equation}
\label{eq: expanded loss 1 (Linear model)}
\begin{split}
    L(\Phi(\vb{a}(0))) &= -\log( \frac{\exp[-E(\vb{a}(T))]}{\int^{\infty}_{-\infty} d\vb{a}\exp[-E(\vb{a})]} ) \\
    &= \log (\int^{\infty}_{-\infty} d\vb{a}\exp[-E(\vb{a})]) + E(\vb{a}(T)) \, ,
\end{split}
\end{equation}
where the first term can be ignored because it is just a constant independent of $\vb{a}(0)$. Then, for instance, if we take $E(\vb{a}) = \vb{a}^2$, the loss function is written as
\begin{equation}
\label{eq: expanded loss 2 (Linear model)}
    L(\Phi(\vb{a}(0))) = \sum_{i=1}^{n} a_i(T)^2 \, ,
\end{equation}
then we can define localized loss function such as
\begin{equation}
\label{eq: localized loss (Linear model)}
    \mathcal{L}_i = a_i(T)^2 \, .
\end{equation}
This model seems good, but it has a major problem in practice. That is, the transformation of $\vb{a}(t)$ is just linear and the model does not have a universal approximation property. In fact, taking $\vb{a}^* (t)= \vb{a}(t) + \tilde{\vb{W}}^{-1} \vb{b}$, we get 
\begin{equation}
\label{eq: linear (Linear model)}
    \vb{a}^* (T) = \exp[\tilde{\vb{W}}T] \vb{a}^* (0) \, ,
\end{equation}
which shows that $\vb{a}^* (T)$ is completely linear to $\vb{a}^* (0)$. To solve this problem, we need to add nonlinearity, and the prescription will be discussed in the next section.

\subsection{Nonlinear model}
\label{sec: Nonlinear model}
\begin{figure}[t]
    \centering
    \includegraphics[scale=0.8]{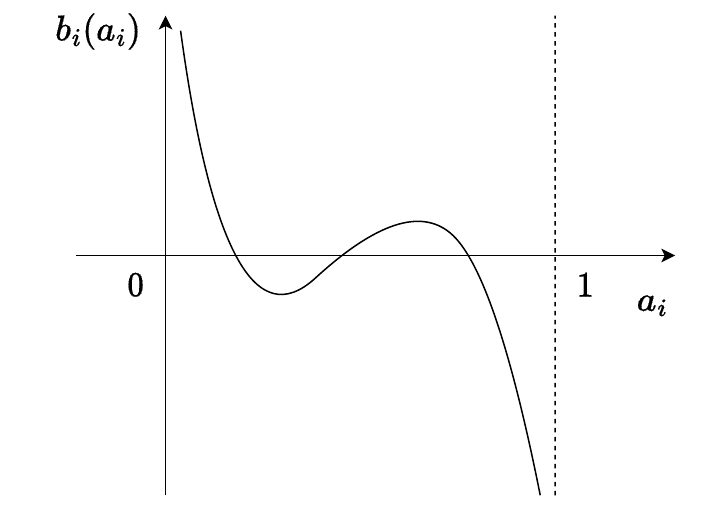}
    \caption{\small \sl The example of the function $b_i(a_i)$.}
    \label{fig: ba}
\end{figure}

Normal machine learning uses an activation function to add nonlinearity, and we consider how to generalize the activation function in the time evolution model. Taking the commonly used Sigmoid function as an example, it is a function defined as $\sigma(x) = (1+\exp(-x))^{-1} $. This function has a role to convert the input from $-\infty$ to $\infty$ into the output from $0$ to $1$. From this observation, we consider that the essence of the activation function is to restrict the range of the output value. In our model, we restrict the value of $\vb{a}(t)$ from 0 to 1, to add nonlinearity. This is done by restricting the initial value $\vb{a}(0)$ from 0 to 1 and modifying the time evolution equation as
\begin{equation}
\label{eq: time evolution (Nonlinear model)}
    a_i(t+dt) - a_i(t) =   \left(\sum_{J=1}^n w_{ij} a_j(t)  + b_i(a_i(t))\right) dt \, ,
\end{equation}
where $b_i(a_i)$ is the continuous function of $a_i$ in $0< a_i < 1$, which satisfies $\lim_{a_i\to 0}b_i(a_i) = \infty$ and $\lim_{a_i\to 1}b_i(a_i) = -\infty$. Fig.\,\ref{fig: ba} shows an example of the function $b_i(a_i)$.  When $a_i$ is close to 1, the value of $b_i$ becomes very small and the value of $a_i$ is suppressed, and when $a_i$ is close to 0, the value of $b_i$ becomes very large and the value of $a_i$ is recovered. As a result, $a_i$ will always remain in the range of 0 to 1. In this model, $b_i$ is no longer a constant, but the shape of the function $b_i(a_i)$ does not change with time. The shape of the function $b_i(a_i)$ itself is the target of optimization.

Although we have added nonlinearity to the time evolution equation, this differential equation is no longer solvable, and the method used in Sec.\,\ref{sec: Linear model} cannot be applied. On the other hand, we can apply the method introduced in Chapter\,\ref{sec: Normalization by differential}, because the range of $\vb{a}(t)$ is $0<\vb{a}(t)<1$. Then we define the model function $\Phi(\vb{a}(0))$ as
\begin{equation}
\label{eq: phi (Nonlinear model)}
    \Phi(\vb{a}(0)) = \frac{\partial\vb{a}(T)}{\partial \vb{a}(0)} \, .
\end{equation}
The normalization condition is easily checked as
\begin{equation}
\label{eq: norm condition (Nonlinear model)}
\begin{split}
    \int_{0}^{1}d\vb{a}(0)  \Phi(\vb{a}(0)) &=  \int_{0}^{1}d\vb{a}(0) \frac{\partial\vb{a}(T)}{\partial \vb{a}(0)} \\
    &= \int_{0}^{1}d\vb{a}(T) \\
    &= 1 \, .
\end{split}    
\end{equation}
Next, we discuss how to calculate Eq.\,(\ref{eq: phi (Nonlinear model)}). As in Fig.\,\ref{fig: allconnected_concept}, the time evolution of this model can be regarded as the layer of a neural network, and we can perform similar calculation as we did in Sec.\,\ref{sec: Remarkable property of the normalization by differential}. We use the chain rule of the Jacobian determinant, and we get 
\begin{equation}
\label{eq: time slice (Nonlinear model)}
    \Phi(\vb{a}(0)) = \frac{\partial\vb{a}(dt)}{\partial \vb{a}(0)}\frac{\partial\vb{a}(2dt)}{\partial \vb{a}(dt)}\cdots \frac{\partial\vb{a}(T)}{\partial \vb{a}(T-dt)}\, .
\end{equation} 
Each Jacobian determinant defined in small time evolution is calculated as
\begin{equation}
\label{eq: jacobian (Nonlinear model)}
\begin{split}
    \frac{\partial\vb{a}(t+dt)}{\partial \vb{a}(t)} &= \left|
\begin{array}{cccc}
 1 + \displaystyle \frac {d b_1(a_1(t))}{d a_1(t)}dt & w_{12}dt & \cdots &  w_{1n}dt  \\
w_{21}dt & 1 + \displaystyle \frac {d b_2(a_2(t))}{d a_2(t)}dt & \cdots & w_{2n}dt \\
\vdots & \vdots &\ddots & \vdots \\
w_{n1} dt & w_{n2} dt & \cdots & 1 + \displaystyle \frac {d b_n(a_n(t))}{d a_n(t)}dt \\
\end{array}
\right| \\
&= \prod_{i=1}^n \left(1+\frac {d b_i(a_i(t))}{d a_i(t)}dt\right) + \mathcal{O}(dt^2) \, ,
\end{split}
\end{equation}
where only the product of diagonal elements survives because other terms contain $dt^2$. Substituting this equation into Eq.\,(\ref{eq: time slice (Nonlinear model)}), we get
\begin{equation}
\label{eq: expanded phi (Nonlinear model)}
    \Phi(\vb{a}(0)) = \prod_{t=0}^{T}\prod_{i=1}^{n}\left(1+\frac {d b_i(a_i(t))}{d a_i(t)}dt\right) \, .
\end{equation}
Then the loss function is calculated as
\begin{equation}
\label{eq: expanded loss (Nonlinear model)}
\begin{split}
    L(\Phi(\vb{a}(0))) &= - \log \left( \prod_{t=0}^{T}\prod_{i=1}^{n}\left(1+\frac {d b_i(a_i(t))}{d a_i(t)}dt\right) \right) \\
    &= - \sum_{i=1}^{n}\int_{0}^{T}dt \, \frac {d b_i(a_i(t))}{d a_i(t)} \, .
\end{split}
\end{equation}
Astonishingly, this loss function is not only localized on each node but also on each time slice. We can define localized loss function as
\begin{equation}
\label{eq: localized loss (Nonlinear model)}
    \mathcal{L}_i(t) =  -\frac {d b_i(a_i(t))}{d a_i(t)} \, .
\end{equation}
The original definition of  $\Phi(\vb{a}(0))$ in this model included values such as $\vb{a}(0)$ and $\vb{a}(T)$ that referred to special times: initial state and final state. However, looking at the loss function, it is written in a localized loss function for each time, so there is no special times in optimization. In other words, it is sufficient to perform optimization sequentially. Furthermore, since the loss function is also localized to the nodes, it is sufficient to optimize the parameters locally without using back propagation.

\subsection{Consideration about the normalization by time evolution}
\label{sec: Remarkable property of the normalization by time evolution}
This model was constructed with reference to the structure of the brain. In this section, we will deepen our understanding by comparing it with the actual brain.

First, let us consider the hardware aspect of the actual brain. The brain is an object composed of proteins, and should only work according to physical laws. In physics, it is known that physical laws are described in terms of local interaction, and any object is affected only locally. Neurons in the brain are no exception to this, and should be affected only by local interactions. That is, optimization of the neurons must also be performed locally, which means that the loss function should be defined locally. Localization of the loss function is required by physical laws, and our model is in line with this.

Next, let us consider the software aspect of the actual brain. The most remarkable property of the brain is its general-purpose learning. It is possible to learn any concept such as causal relationships, for instance ``A happens because B happened.'' The human brain is constantly receiving information from the five senses, and it is considered to find causal relationships by sequentially processing such temporally continuous information. The fact that information is processed sequentially means that optimization is performed at each time slice, and this property is common to this model. In other words, this model can also learn concepts such as causal relationships. For example, if we input video data into our model as multiple image data sequentially in time, it will automatically acquire these concepts.

With these in mind, this model is considered as a mathematical realization of the learning mechanism in the brain. However, there are two major problems remaining. The first is, if this model is correct mathematically, how it is actually realized in the brain. Neurons in the brain contain various substances and electrical signals, and we need to reveal how these are related to the variables in our model. The second is how to put this model into practical. We still do not have the knowledge necessary for concrete implementation, such as how many nodes are sufficient, how much time resolution is required for time evolution, and what is an efficient way to optimize parameters. If these issues are resolved, this model can become an artificial general intelligence that can learn from any data, just like humans.

\section{Discussion}
\label{sec: Discussion}
In this chapter, we will consider the meaning of the learning principle and the interpretations derived from it. First, we will explain the concept of features, which are considered important in conventional machine learning, from the perspective of learning principle. Based on the learning principle, all information is included in the probability distribution of the input dataset. In particular, the concept of feature corresponds to the part where the probability is maximum in a certain phase space. We will explain it using an example. When considering the task of handwritten digit recognition, each digit is made up of features such as how lines are drawn. For example, if you represent the curved part of the number ``3'' with a straight line or separate the top and bottom, it will be difficult to recognize it as ``3''. This is due to the fact that while numbers drawn in their normal form appear frequently, numbers drawn in distorted forms seldom appear. In other words, something regarded as a feature has a higher probability of appearance than distorted versions of it. The important thing is that although there are a large number of the way to distort the shape, the probability of appearance is concentrated only in a very small number of normal patterns. This probability bias is the essence of features.

Next, we will explain why deep learning has been successful. According to the learning principle, as long as the model function satisfies the normalization condition, it automatically approaches the true probability. In particular, if a model has a universal approximation property and there are infinite datasets and computational resources, it can get as close to the true probability distribution as possible. Namely, the success of deep learning is due to achieving universal approximation property by increasing the depth of the layers, and the recent improvements in computer processing power. It is sometimes explained that deep learning is successful because it can capture complex features, but the concept of feature is merely an afterthought that emerges when probability distributions are successfully estimated.

Now, we will explain what it means to improve a model in deep learning. According to the learning principle, as long as the model has a universal approximation property, the same solution will eventually be obtained regardless of the structure of the model. What changes depending on the structure of the model is the speed of convergence to the solution. For example, if the characteristics that image data or language data are incorporated into the model structure in advance, the speed of convergence to the solution will become overwhelmingly faster for such specific data. In reality, since there are no infinite datasets or computational resources, learning must be stopped midway through, and as a result, models that converge quickly achieve good results. This causes a misunderstanding that the essence of learning is hidden in the structure of the model. Improving a model means making it specialized for specific inputs.

Finally, we mention two methods that we devised. Both methods are built on the principle of learning, and satisfy normalization conditions without any assumptions or approximations. This shows that they are general-purpose models that can be used without specifying the type of input data. In fact, the results in Figs.\,\ref{fig: demonstration1},\ref{fig: demonstration2},\ref{fig: demonstration3} demonstrate the versatility of the method using differentiation. We also showed that a method considering the time evolution of a fully or partially connected model can be regarded as a mathematical realization of the learning mechanism in the brain. These accomplishments are consequences of the learning principle, and on the contrary, they guarantee the correctness of the learning principle.

\section{Conclusion}
\label{sec: Conclusion}
In this paper, we derived a learning principle that uniformly describes all learning, including machine learning and learning in the brain. Under this principle, all learning is understood as a probability estimation of input data. The conditions for applying this principle are that the estimated probability is always positive and satisfies the normalization condition, and that optimization is performed using a loss function defined by the logarithm of the estimated probability. Conversely, as long as these conditions are met, anything can be considered as learning. We confirmed that supervised learning also satisfies the learning principle and can be regarded as a type of probability estimation of conditional probabilities.

In the learning principle, the loss function is always defined as the logarithm of the model function with a negative sign, and what can be changed is the method to satisfy the normalization conditions of the model function. In this paper, we proposed two methods that satisfy the normalization condition without any assumptions or approximations. Both methods are considered to be general-purpose methods that can perform learning with high accuracy on any dataset. The first method uses differentiation to satisfy the normalization condition, and was actually trained on several datasets, to exhibit very good behavior. The second method is based on the structure of neurons and synapses in the brain, and satisfies the normalization condition by considering time evolution in the fully or partially connected model. Furthermore, we showed that the loss function defined there can be expressed as a sum of functions localized in space and time. This means that parameter optimization can be performed sequentially and locally without using back propagation. Even in the actual brain, optimization must be performed sequentially and locally as a result of the laws of physics. Considering these facts, this method can be regarded as a mathematical realization of the learning mechanism in the brain. This will be a major stepping stone towards the realization of artificial general intelligence.

Furthermore, we reviewed conventional machine learning from the perspective of learning principle and provided a new understanding. In learning, all information is included in the input probability distribution, and concepts such as features are also defined based on probability. Improving the model corresponds to speeding up the convergence to a solution by making it specialized for a particular data set. In aiming to generalize machine learning, it is important to understand learning itself based on the learning principle, rather than ad-hoc measures.

\section*{Acknowledgments}
The author would like to thank H.\,Katayose, H.\,Tsuji and N.\,Okada for constructive comments.

\bibliographystyle{apsrev4-1}
\bibliography{refs}

\end{document}